\documentclass[twocolumn, switch]{article} 

\usepackage{preprint}
\usepackage{algorithm}
\usepackage{algpseudocode}
\usepackage{amsmath, amsthm, amssymb, amsfonts}

\usepackage[numbers,square]{natbib}
\usepackage{diagbox} 
\usepackage{amsmath}
\usepackage{amssymb}   
\usepackage{multirow}
\usepackage{booktabs}
\usepackage{enumitem}

\usepackage[utf8]{inputenc}	
\usepackage[T1]{fontenc}	
\usepackage{xcolor}		
\usepackage[colorlinks = true,
            linkcolor = purple,
            urlcolor  = blue,
            citecolor = cyan,
            anchorcolor = black]{hyperref}	
\usepackage{booktabs} 		
\usepackage{nicefrac}		
\usepackage{microtype}		
\usepackage{lineno}		
\usepackage{float}			

\usepackage{lipsum}		

\usepackage{newfloat}
\DeclareFloatingEnvironment[name={Supplementary Figure}]{suppfigure}
\usepackage{sidecap}
\sidecaptionvpos{figure}{c}

\usepackage{titlesec}
\titlespacing\section{0pt}{12pt plus 3pt minus 3pt}{1pt plus 1pt minus 1pt}
\titlespacing\subsection{0pt}{10pt plus 3pt minus 3pt}{1pt plus 1pt minus 1pt}
\titlespacing\subsubsection{0pt}{8pt plus 3pt minus 3pt}{1pt plus 1pt minus 1pt}

\title{SculptDrug : A Spatial Condition-Aware Bayesian Flow Model for Structure-based Drug Design}

\usepackage{eso-pic}
\usepackage{tikz}
\usepackage{xcolor}
\PassOptionsToPackage{colorlinks=true,linkcolor=gray,urlcolor=gray}{hyperref}

\newcommand{\AddMyWatermarks}{%
  \begin{tikzpicture}[remember picture, overlay]
    \node[rotate=90, color=gray!60, scale=1] at ([xshift=-4.05in,yshift=0in]current page.center) {%
    };
    \node[rotate=90, color=gray!60, scale=1] at ([xshift=3.9in,yshift=0in]current page.center) {%
    };
    \node[color=gray!90, scale=1] at ([xshift=0in,yshift=-5in]current page.center) {%
      This is the author's accepted manuscript (extended version).. The final version will appear in AAAI 2026, © AAAI 2026.%
    };
  \end{tikzpicture}%
}

\AddToShipoutPictureBG*{\AddMyWatermarks}

\usepackage{titling}
\usepackage{orcidlink}
\usepackage{footmisc}
\author{
  Qingsong Zhong$^{1}$,
  Haomin Yu$^{3}$,
  Yan Lin$^{4}$,
  Wangmeng Shen$^{1}$,
  Long Zeng$^{1}$,
  Jilin Hu$^{1,2}$\thanks{Corresponding author. Email: jlhu@dase.ecnu.edu.cn} \\[0.5em]
  \small $^{1}$School of Data Science and Engineering, East China Normal University, China\\
  \small $^{2}$School of Chemistry and Molecular Engineering, East China Normal University, China\\
  \small $^{3}$School of Science, Engineering \& Environment, University of Salford, UK\\
  \small $^{4}$Department of Computer Science, Aalborg University, Denmark\\[0.25em]
  \small \texttt{\{xxrelax, wmshen, longzeng\}@stu.ecnu.edu.cn, h.yu6@salford.ac.uk, lyan@cs.aau.dk, jlhu@dase.ecnu.edu.cn}
}

\begin{document}

\twocolumn[ 
  \begin{@twocolumnfalse} 

\maketitle
\thispagestyle{empty}

\begin{abstract}
Structure-Based Drug Design (SBDD) has emerged as a popular approach in drug discovery, leveraging three-dimensional protein structures to generate drug ligands. However, existing generative models encounter several key challenges: (1) Incorporating boundary condition constraints, (2) Integrating hierarchical structural conditions and (3) Ensuring spatial modeling fidelity. To overcome these limitations, we propose SculptDrug, a spatial condition-aware generative model based on Bayesian Flow Networks (BFNs). 
First, SculptDrug  follows a BFNs-based framework and employs a progressive denoising strategy to ensure spatial modeling fidelity, iteratively refining atom positions while enhancing local interactions for precise spatial alignment.
Second, we introduce the Boundary Awareness Block, which incorporates protein surface constraints into the generative process to ensure that the generated ligands are geometrically compatible with the target protein.
Finally, we design a Hierarchical Encoder that captures global structural context while preserving fine-grained molecular interactions, ensuring overall consistency and accurate ligand–protein conformations.
We evaluate SculptDrug  on the CrossDocked dataset, and experimental results demonstrate that SculptDrug  outperforms state-of-the-art baselines, proving the efficacy of spatial condition-aware modeling. Our implementation is available at \href{https://github.com/decisionintelligence/SculptDrug.git}{https://github.com/decisionintelligence/SculptDrug.git}.
\end{abstract}
\vspace{0.35cm}

  \end{@twocolumnfalse} 
] 



\section{Introduction}
Drug discovery is the process of identifying potential therapeutic molecules to treat diseases, playing a crucial role in improving health and addressing unmet medical needs~\cite{segler2018generating}. Yet, this process is highly complex and time-consuming. Structure-Based Drug Design~(SBDD) has emerged as a powerful approach to streamline this process~\cite{anderson2003process}, leveraging the structural information of biological targets to rationally design molecules with improved specificity and efficacy~\cite{isert2023structure}. 

\begin{figure}[t]
  \centering
  \includegraphics[width=0.90\linewidth]{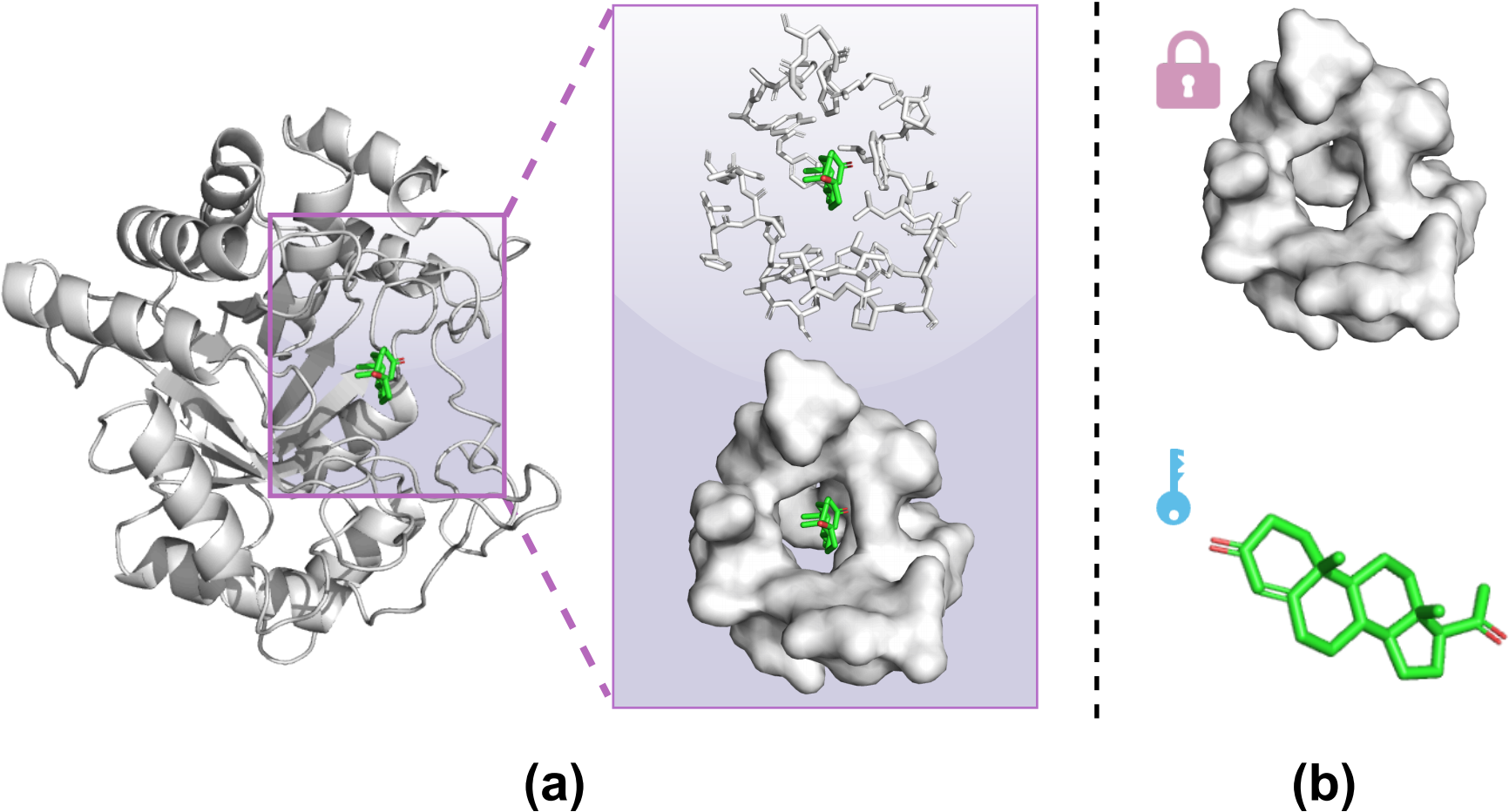}
  \caption{Protein-ligand representation: (a) The protein ribbon model highlights the binding pocket, with close-ups showing the pocket in stick (top) and surface (bottom) representations, emphasizing spatial complementarity. (b) A symbolic ``lock-and-key'' analogy illustrates the specificity of protein-ligand binding.}
  \label{fig:1}
\end{figure}

The efficacy of SBDD lies in its ability to leverage the lock-and-key principle of protein-ligand interactions~\cite{jorgensen1991rusting}, providing a framework for designing molecules that precisely target biological systems, as shown in Figure~\ref{fig:1}.
 Specifically, the protein surface functions as a unique ``lock'', while the ligand serves as the complementary ``key''. This structural compatibility ensures precise protein-ligand interactions, highlighting the effectiveness of SBDD in developing targeted and efficient drugs.
Traditionally, SBDD relies on virtual screening, which aims to select ligands from an immense chemical space estimated to contain approximately $10^{60}$ potential molecules~\cite{reymond2012enumeration}. virtual screening focuses on predefined compound libraries based on molecular properties and structural features~\cite{lionta2014structure}, which cover only a small fraction of the space and significantly limit ligand discovery~\cite{reymond2012enumeration}. Recent deep generative models~\cite{kingma2013auto,goodfellow2020generative,ho2020denoising,bfns} enable ligand generation beyond predefined libraries, supporting broader exploration of chemical space~\cite{cheng2021molecular}.

\begin{figure}[t]
  \centering  \includegraphics[width=0.90\linewidth]{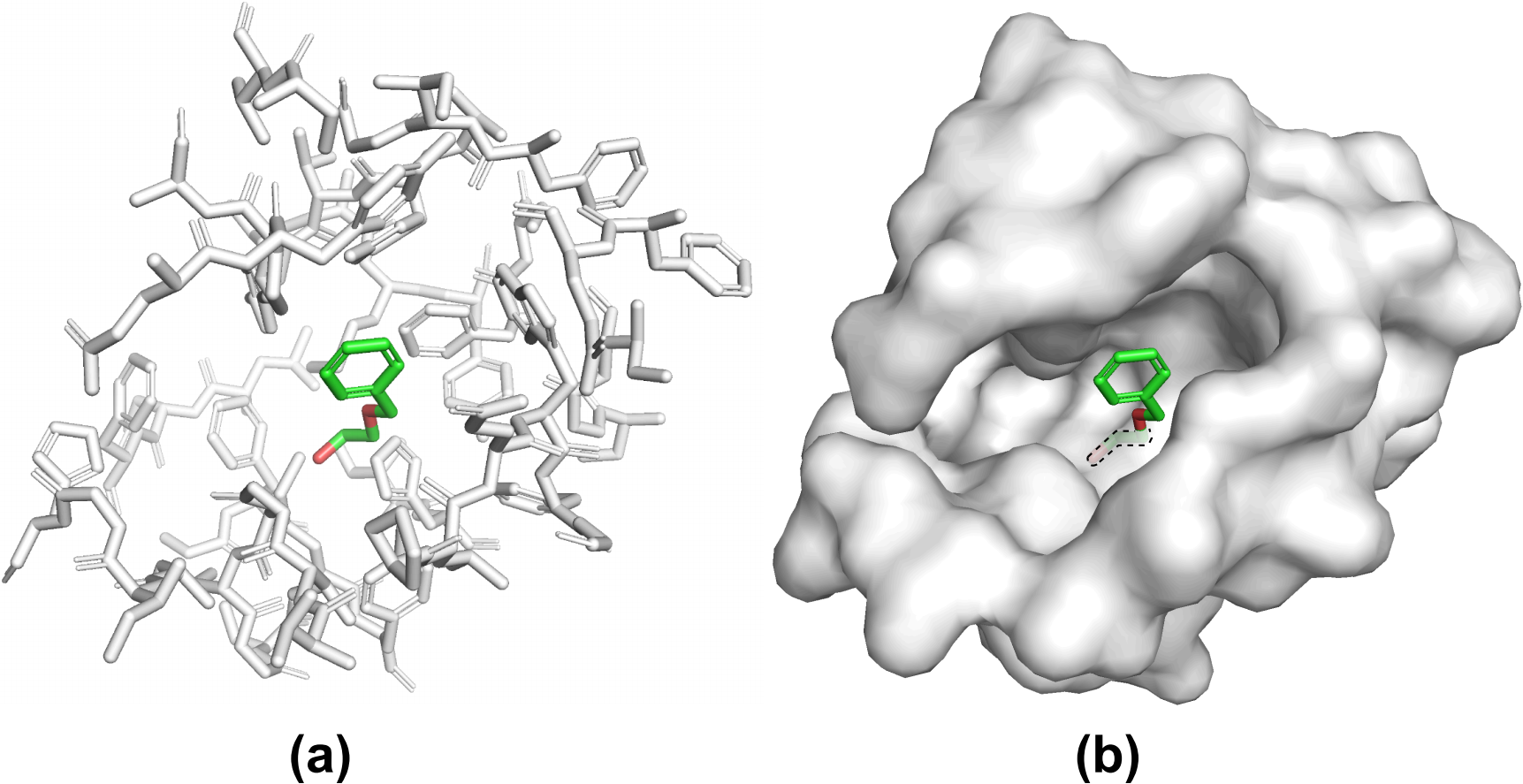}
  \caption{
(a) The generated ligand exhibits reasonable atomic distances when evaluated against protein atoms.  
(b) However, it erroneously penetrates the solvent-excluded surface, violating spatial plausibility.
}
  \label{fig:2}
\end{figure}

However, existing generative models still face some challenges in learning patterns from protein-ligand complexes. These challenges can be categorized into the following aspects. 
(1) \textit{Boundary Condition Constraints.} Designing effective ligands is challenging due to boundary conditions, as ligands are often encapsulated by the protein surface. Due to the lock-and-key nature of protein-ligand interactions, effective ligands must fit within the protein surface boundary. Existing generative models often ignore these constraints, leading to misaligned or buried ligands~(Figure~\ref{fig:2}).
(2)\textit{~Hierarchical Structure Condition Integration.} Integrating hierarchical structural conditions remains a challenge in ligand design. The local structure is crucial for determining the precise alignment of the key's teeth within a lock, ensuring a better fit based on the lock-and-key principle. Meanwhile, the global structure aims to capture the overall shape of the lock, offering a broader perspective, as the lock’s shape and integrity guide the key’s design to achieve proper alignment and effective functionality. It is essential to seamlessly integrate both local and global structural information to design ligands that are both functionally effective and structurally stable.

Diffusion models and Bayesian Flow Networks (BFNs)~\cite{bfns} are powerful tools for molecular generation, but often struggle with (3)~\textit{Spatial Modeling Fidelity}. Unlike images with fixed spatial layouts, molecules rely on flexible interatomic distances to define structure. Noise added to atomic coordinates can distort these distances, causing atoms to fall outside interaction thresholds and be excluded from local computations, ultimately compromising chemical validity.
 
To address the above challenges, we propose SculptDrug, a spatial condition-aware generative model based on BFNs, designed to generate geometrically accurate and chemically plausible drug-like ligands.
First, to improve spatial modeling fidelity, SculptDrug adopts a progressive denoising strategy that gradually refines atom positions and types, enabling more accurate modeling of protein–ligand interactions.
Second, we introduce the Boundary Awareness Block, which encodes protein surface information to guide ligand placement and ensure geometric compatibility, effectively avoiding steric clashes.
Finally, we incorporate a Hierarchical Encoder that integrates both global structural constraints, ensuring the ligand fits the overall protein pocket, and fine-grained local interactions, which resemble the alignment of a key’s teeth within a lock. This design enables the model to maintain structural context integrity while effectively capturing conformational complexity across multiple structural levels.


Overall, the main contributions of this work are summarized as follows: 
\begin{itemize} 
    \item We propose a structure-based drug design framework, SculptDrug, which leverages a progressive denoising strategy to achieve high-fidelity spatial interaction modeling. 
    \item We introduce the Boundary Awareness Block to incorporate protein surface geometry into the generative process, encouraging the generated ligands to align with structural constraints and reducing the likelihood of steric clashes.
    \item We design a Hierarchical Encoder that incorporates both global and local structural contexts, supporting multi-scale alignment.
    \item We validate the effectiveness of SculptDrug  through extensive experiments, demonstrating its superior performance in generating drug-like ligands compared to existing methods. 
\end{itemize}

\section{Related Work}
Ligand generation has evolved from 1D representations like SMILES~\cite{bjerrum2017molecular,segler2018generating}, to 2D molecular graphs~\cite{liu2018constrained,jin2018junction}, and more recently to 3D structure-based approaches. While 1D/2D methods capture some chemical constraints, they lack the spatial detail essential for modeling protein-ligand interactions. With the rise of deep learning in 3D modeling, structure-aware ligand generation has become a central research focus. Existing methods can be broadly categorized into: (1)~\textit{Voxel-based} and (2)~\textit{Euclidean space-based} models.

\textit{Voxel-based methods} discretize 3D space into regular grids and model ligand density distributions within this voxelized space. Representative works such as LIGAN~\cite{ligan} and VoxBind~\cite{voxbind} generate ligands aligned to protein binding pockets using voxel-based representations. However, these methods suffer from high computational cost at fine resolutions and limited spatial precision due to the loss of structural detail during voxelization.

\textit{Euclidean space-based methods} generate atomic types and positions directly in continuous 3D coordinates, typically representing proteins and ligands as point clouds or graphs. Early models, including GraphBP~\cite{Graphbp}, Pocket2Mol~\cite{pocketmol}, and FLAG~\cite{Flag}, adopt autoregressive strategies at either the atom or fragment level. SurfGen~\cite{surfgen} and ResGen~\cite{resgen} incorporate protein context through surface features or residue-level graphs. However, the autoregressive nature of these methods can lead to error accumulation and limited global awareness.

Diffusion-based approaches, such as DIFFBP~\cite{Diffbp}, TARGETDIFF~\cite{targetdiff}, and DIFF\-SBDD~\cite{diffsbdd}, leverage 3D-equivariant denoising to iteratively refine atom placements, improving spatial consistency and structural integrity. Extensions like D3FG~\cite{D3fg} and DECOMPDIFF~\cite{Decompdiff} introduce fragment priors and scaffold decomposition to enhance diversity and controllability. Nonetheless, modeling discrete atom types alongside continuous coordinates remains challenging. The introduction of Bayesian Flow Networks adds a new dimension to molecular generation ~\cite{GeoBFN}. MOLCRAFT~\cite{molcraft} generates ligands in continuous parameter spaces. However, their limited ability to perceive protein spatial geometry constrains performance in structure-guided ligand design.
\begin{figure*}[ht]  
	\centering  
	\includegraphics[width=1\textwidth]{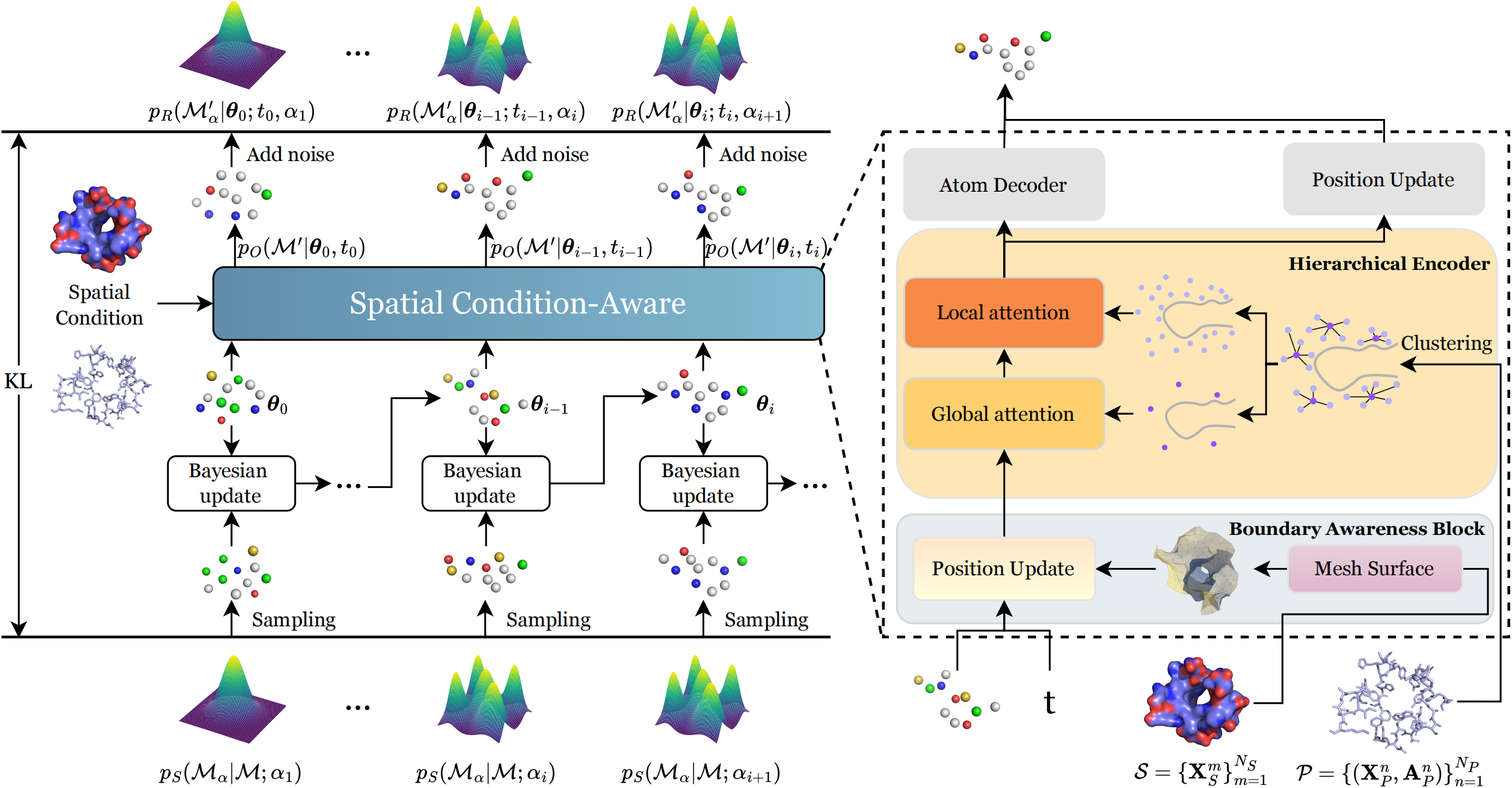}
	\caption{Overview of the SculptDrug framework for ligand generation.}
	\label{fig1}  
\end{figure*}
\section{Preliminaries}
\subsection{Problem Definition}
\textit{Definition 1 (Protein structure)}. The protein structure refers to the three-dimensional arrangement of atoms within a protein molecule. It is represented as \(\mathcal{P} = \{ (\mathbf{x}_p^n, \mathbf{a}_p^n) \}_{n=1}^{N_p},\) where \( \mathbf{x}_p^n \) denotes the 3D coordinates of the \( n \)-th atom in the protein, \( \mathbf{a}_p^n \) represents the atom type, and \( N_P \) is the total number of atoms in the protein.

\textit{Definition 2 (Protein surface).}  
The protein surface refers to the solvent-excluded surface that defines the geometric boundary of a protein. It is represented as a surface graph  
\(
\mathcal{S} = \left( \left\{ (\mathbf{x}_s^n, \mathbf{a}_s^n) \right\}_{n=1}^{N_s}, \mathcal{E}_s \right),
\)
where \( \mathbf{x}_s^n \in \mathbb{R}^3 \) is the coordinate of the \( n \)-th surface vertex, \( \mathbf{a}_s^n \in \mathbb{R}^d \) is the associated feature vector describing geometric and biochemical properties (e.g., shape index, hydrophobicity, polarity, electrostatic charge), \( N_s \) is the number of surface points, and \( \mathcal{E}_s \subseteq \{(i,j)\} \) defines edges connecting nearby surface points.

\textit{Definition 3 (Ligand)}. The generated ligand molecule, denoted as 
\(
\mathcal{M} = \{ (\mathbf{x}_m^n, \mathbf{a}_m^n) \}_{n=1}^{N_m},
\)
consists of atom position \( \mathbf{x}_m^n \) and type \( \mathbf{a}_m^n \), where \( N_m \) represents the total number of atoms in the ligand.

The task can be formally defined as follows: Given the protein structure \( \mathcal{P} \) and its surface representation \( \mathcal{S} \), the goal is to generate a ligand molecule \( \mathcal{M} \) that exhibits high binding affinity, favorable drug-likeness, and a well-formed 3D conformation. This process can be expressed as:
\begin{equation}
\mathcal{F}(\mathcal{S}, \mathcal{P}) = \mathcal{M},
\end{equation}
where \( \mathcal{F} \) denotes the generative function that maps the protein structure and surface to a chemically and geometrically plausible ligand.

\subsection{Bayesian Flow Networks}
Bayesian Flow Networks (BFNs) represent a novel generative modeling framework that combines Bayesian inference with neural networks to iteratively refine distribution parameters for generating complex data distributions.

In each iteration, BFNs define a sender distribution $p_S(\mathbf{y}|\mathbf{x}, \alpha(t))$ that describes how the noisy observations $\mathbf{y}$ arises from data $\mathbf{x}$ under noise level $\alpha(t)$. The sender distribution can be expressed as:
\begin{equation}
p_S(\mathbf{y}|\mathbf{x}, \alpha(t)) = \prod_{d=1}^D p_S(y^{(d)}|x^{(d)}; \alpha(t)),
\end{equation}

Through Bayesian updating, we iteratively refine the parameter \(\boldsymbol{\theta}\) based on noisy observations \(\mathbf{y}\). Over time, the randomness introduced by the noise causes \(\boldsymbol{\theta}\) to evolve stochastically. By applying multiple Bayesian updates, we eventually obtain the Bayesian flow distribution, which represents the marginal distribution of the parameters $\boldsymbol{\theta}$ over all updates up to time step $t$. It is defined as:
\begin{equation}
p_F(\boldsymbol{\theta}|\mathbf{x}; t) = p_U(\boldsymbol{\theta}|\boldsymbol{\theta}_0, \mathbf{x}; \beta(t)),
\end{equation}
where $\beta(t)$ represents the total noise intensity over all previous updates.

Bayesian updates only perform independent inference for each dimension, and BFNs leverage neural networks to incorporate contextual information, deriving the output distribution $p_O(\mathbf{x}|\boldsymbol{\theta}; t)$ by feeding the Bayesian-updated parameter $\theta$ and the time step $t$ into the network for more accurate predictions.

The loss function in BFNs aims to minimize the divergence between the sender distribution $p_S(\mathbf{y}|\mathbf{x}, \alpha(t))$ and the receiver distribution $p_R(\mathbf{y}|\boldsymbol{\theta}; t, \alpha(t))$. Thus, the total loss $\mathcal{L}(\mathbf{x})$ consists of two components:
$n$-step loss $\mathcal{L}_n(\mathbf{x})$ and reconstruction loss \begin{align}
\mathcal{L}(\mathbf{x}) 
    &= \mathcal{L}_n(\mathbf{x}) + \mathcal{L}_r(\mathbf{x}), \\
\mathcal{L}_n(\mathbf{x}) 
    &= \mathbb{E}_{p(\boldsymbol{\theta}_1, \ldots,\boldsymbol{\theta}_{n-1})}
       \sum_{i=1}^n D_{KL}\bigl(p_S^i \,\|\, p_R^i\bigr), \\
\mathcal{L}_r(\mathbf{x}) 
    &= - \mathbb{E}_{p_F(\boldsymbol{\theta}|\mathbf{x}, 1)} 
       \ln p_O(\mathbf{x}|\boldsymbol{\theta}; 1).
\end{align}

By introducing sender and receiver distributions and minimizing their divergence, BFNs provide a flexible and effective framework for generative modeling.

\section{Spatial Condition-Aware Model}
To advance SBDD, we propose a novel model, SculptDrug , based on BFNs. SculptDrug  enhances the ligand generation process by integrating both the surface and structural information of the target protein. At each step, the model progressively captures the conditional information of the protein, refining it from coarse to fine, ensuring that the generated ligands adhere to fundamental biological and chemical principles.

As shown in Figure \ref{fig1}, during the BFNs process, at the initial step (\(i=0\)), a prior input distribution is provided. In each subsequent transmission step \(i\), the parameters \(\boldsymbol{\theta_{i-1}}\) of the previously learned distribution are input into SculptDrug . The model then reconstructs the ligand structure $\mathcal{M'}$ before noise is added and generates an output distribution through a Spatial Condition-Aware (SCA) neural network. The output distribution is given by:
\begin{equation}
p_O(\mathcal{M'} \mid \boldsymbol{\theta}_{i-1}, t_{i-1})
= \text{SCA}(\mathcal{S}, \mathcal{P}, \boldsymbol{\theta}_{i-1}, t_{i-1}).
\end{equation}

Subsequently, the sender modifies the precision of the ligand data based on a predefined schedule to obtain the sender distribution \(p_S(\mathcal{M}_\alpha|\mathcal{M}; \alpha_i)\). Simultaneously, the receiver distribution \(p_R(\mathcal{M}_\alpha'|\boldsymbol{\theta}_{i-1}; t_{i-1}, \alpha_{i})\) is calculated by applying the same precision to the output distribution. A sample is then drawn from the sender distribution, and through Bayesian updates, the input distribution parameters \(\boldsymbol{\theta}_i\) are refined for the next round of transmission. After multiple iterations, the data's distribution progressively transitions from the prior distribution to a posterior distribution that more accurately approximates the true distribution of ligand structure.

Importantly, Bayesian updates are performed independently for each data dimension, while the Spatial Condition-Aware neural network plays a crucial role in integrating contextual and conditional information across dimensions, ensuring a more accurate and coherent reconstruction of the ligand structure.

To enhance the reconstruction of denoised ligands under spatial and conditional constraints, we introduce two key modules. The \textbf{Boundary Awareness Block} incorporates protein surface information to guide ligand placement within chemically and structurally plausible regions. The \textbf{Hierarchical Encoder} captures multi-scale protein context by jointly modeling global pocket geometry and local atomic interactions, enabling precise structural conditioning during ligand generation.


\subsection{Boundary Awareness Block}

In drug design, the binding of ligand to target proteins relies heavily on their precise spatial matching. This relationship is analogous to a key fitting into a lock, where the ligand, as the ``key'', must align perfectly with the ``lock'' represented by the protein. To address this, we propose the Boundary Awareness Block, which enhances the model’s ability to understand the geometry of protein surfaces and incorporate this essential spatial information during ligand generation.

To achieve this, we follow a systematic approach for understanding and simplifying the protein ``lock'' and encoding the ligand ``key''.
First, we decode the lock by extracting and simplifying the protein surface structure. Once the surface geometry and spatial features of the lock are identified, we encode the ligand key, ensuring both components are optimized for precise spatial matching. The final stage involves adapting the ``key'' to fit the ``lock'' by updating the atomic positions of the ligand.

\subsubsection{Extracting the Protein Surface.}  
Inspired by SurfGen~\cite{surfgen} and SurfPro~\cite{song2024surfpro}, we extract the surface structure of the binding pocket by selecting residues within 10\,\AA{} of any ligand atom. The solvent-excluded surface is computed using MSMS~\cite{msms}, which outputs a triangular mesh of surface vertices and faces. We retain only the inward-facing vertices and their adjacent edges that are spatially close to the ligand. Each vertex is annotated with geometric and biochemical features: the shape index is computed from local curvature, while hydrophobicity, polarity, and electrostatic charge are assigned based on the nearest residue.
The resulting surface graph aligns with the formal definition of \( \mathcal{S} \).

\subsubsection{Fitting the Key to the Lock.}  

We construct a unified spatial graph \( \mathcal{G} = (\mathcal{V}, \mathcal{E}) \) by merging protein surface points and ligand atoms. The edge set \( \mathcal{E} \) consists of both the original mesh connectivity \( \mathcal{E}_S \) from the protein surface and additional \( k \)-nearest neighbor (k-NN) edges constructed in Euclidean space. 
Each edge \( (i, j) \in \mathcal{E} \) is annotated with a 6-dimensional one-hot vector \( \mathbf{t}_{ij} \in \mathbb{R}^6 \), indicating both the types of the connected nodes and the source of the edge. The Euclidean distance between nodes is further encoded using Gaussian radial basis functions:
\begin{align}
\boldsymbol{\phi}_{ij} = \phi(\|\mathbf{x}_i - \mathbf{x}_j\|) \in \mathbb{R}^g,
\end{align}
We construct edge features via outer product:
\begin{align}
\mathbf{e}_{ij} = \mathbf{t}_{ij} \otimes \boldsymbol{\phi}_{ij} \in \mathbb{R}^{6 \times g},
\end{align}
which is then flattened and concatenated with the source and target node features:
\begin{align}
\tilde{\mathbf{e}}_{ij} = \text{concat}(\mathbf{h}_i, \mathbf{h}_j, \text{flatten}(\mathbf{e}_{ij})) \in \mathbb{R}^{2d + 6g}.
\end{align}

Given these representations, the attention weights and vector messages are defined as:
\begin{align}
\alpha_{mj} &= \text{softmax}_j\left( \frac{f_q(\mathbf{h}_m)^\top f_k(\tilde{\mathbf{e}}_{mj})}{\sqrt{d}} \right), \\
\mathbf{v}_{mj} &= f_v(\tilde{\mathbf{e}}_{mj}) \cdot (\mathbf{x}_j - \mathbf{x}_m),
\end{align}
where \( f_q, f_k, f_v \) are learnable multi-layer perceptrons (MLPs), and \( d \) denotes the hidden dimension.

The ligand atom positions \( \mathbf{x}_m \) are updated via message aggregation:
\[
\mathbf{x}_m \leftarrow \mathbf{x}_m + \Delta \mathbf{x}_m, \quad \text{where} \quad \Delta \mathbf{x}_m = \sum_{j \in \mathcal{N}(m)} \alpha_{mj} \cdot \mathbf{v}_{mj},
\]
with \( \mathcal{N}(m) \) denoting the spatial neighbors of node \( m \) in graph \( \mathcal{G} \).

\subsection{Hierarchical Encoder}

To better balance the local and global structural information of the protein, we propose a layered encoder that extracts features from multiple structural levels, enhancing the model’s understanding of protein context during ligand generation.

While atomic point clouds provide fine-grained local details, they often fail to capture higher-order structural patterns and are sensitive to noise. To address this, we first generate \textit{virtual atoms} by clustering protein atoms, producing a coarse-grained representation that aggregates within-cluster features and highlights global structure.
Ligands then interact with these virtual atoms to integrate global context, while fine-grained interactions are captured via multi-type edges that model detailed biochemical cues.
To support both levels of interaction, we adopt \textit{equivariant attention mechanisms} in the global and local modules, drawing inspiration from the attention design in DecompDiff~\cite{Decompdiff}.

\subsubsection{Generating Virtual Atoms for Protein Representation.}
To obtain a coarse-grained protein representation, we apply k-means++~\cite{arthur2006k} clustering to protein atoms based on 3D coordinates, generating virtual atoms at cluster centroids. Let \( C \) denote the atoms in a cluster. The virtual atom's position is defined as:

\begin{equation}
\mathbf{x}_v = \frac{1}{|C|} \sum_{i \in C} \mathbf{x}_i^P.
\end{equation}

To compute its feature, we perform a distance-aware aggregation from all atoms in the cluster to the virtual atom:

\begin{equation}
\mathbf{h}_v 
= \sum_{i \in C(v)} \alpha_{iv}\,
  \mathrm{MLP}\Bigl([\mathbf{h}_i^P,\; \phi(\|\mathbf{x}_i^P - \mathbf{x}_v\|)]\Bigr),
\end{equation}

where $\boldsymbol{\phi}(\cdot)$ is a radial basis expansion of the interatomic distance, 
and $\alpha_{iv} \in \mathbb{R}$ is a learned scalar aggregation weight produced by an MLP over the same input, 
normalized with a softmax over atoms $i \in C(v)$.
. Unlike individual atoms, each virtual atom encodes a higher-level abstraction of local structure, offering a coarser yet semantically rich representation. This abstraction allows the global attention mechanism to operate over a reduced node set, significantly improving efficiency while retaining key spatial information.




\begin{table*}[t]
    \centering
    \setlength{\tabcolsep}{3pt}

    \begin{tabular}{lccccccccccccccc}
    \toprule
    \multirow{2}{*}{\diagbox{\textbf{Methods}}{\textbf{Metrics}}} 
    & \multicolumn{3}{c}{\textbf{Vina Score}}  
    & \multicolumn{1}{c@{\hspace{1mm}}}{}
    & \multicolumn{3}{c}{\textbf{Vina Min}}
    & \multicolumn{1}{c@{\hspace{1mm}}}{}
    & \multicolumn{3}{c}{\textbf{Vina Dock}}
    & \multicolumn{1}{c@{\hspace{1mm}}}{}
    & \multicolumn{2}{c}{\textbf{Drug-Likeness}} 
    \\
    \cmidrule(lr){2-4} \cmidrule(lr){6-8} \cmidrule(lr){10-12} \cmidrule(lr){14-15}
    & Evina & IMP\% & MPBG\%
    & & Evina & IMP\% & MPBG\%
    & & Evina & IMP\% & MPBG\%
    & & QED & SA 
    \\
    \midrule
    GRAPHBP (2022)  
    & --    & 0.00     & --   
    & & --    & 1.67  & -- 
    & & -4.57 & 10.86 & -30.03 
    & & 0.44 & 0.64 
    \\
    POCKET2MOL (2022)
    & -5.23 & 31.06 & -15.03   
    & & -6.03 & 38.04 & -4.95   
    & & -7.05 & 48.07 & -0.17  
    & & 0.39 & 0.65
    \\
    TARGETDIFF (2023)
    & -5.71 & 38.21 & -22.80 
    & & -6.43 & 47.09 & -1.60 
    & & -7.41 & 51.99 & 5.38  
    & & \underline{0.49} & 0.60  
    \\
    FLAG (2023)    
    & --    & 0.04  & --
    & & --    & 3.44  & --    
    & & -3.65 & 11.78 & -47.64
    & & 0.41 & 0.58
    \\
    D3FG (2023)   
    & --    & 3.70  & --
    & & -2.59 & 11.13 & -67.37
    & & -6.78 & 28.90 & -8.85 
    & & \underline{0.49} & \underline{0.66}
    \\
    DECOMPDIFF (2023)
    & -5.18 & 19.66 & -17.17    
    & & -6.04 & 34.84 & -6.78    
    & & -7.10 & 48.31 & -1.59 
    & & \underline{0.49} & \underline{0.66}
    \\
    DIFFSBDD (2024)
    & --    & 12.67 & -- 
    & &-2.15 & 22.24 & --    
    & & -5.53 & 29.76 & -23.51 
    & & \underline{0.49} & 0.34
    \\
    MOLCRAFT (2024)  
    & \underline{-6.59} & \underline{54.86} & \underline{-1.86} 
    & & \underline{-7.17} & \textbf{61.02} & \underline{10.42}
    & & \underline{-7.83} & \underline{58.05} & \underline{8.17}  
    & & 0.50 & \textbf{0.67}
    \\
    DIFFBP (2025)   
    & --    & 8.60   & --    
    & & --   & 19.68 & --    
    & & -7.34 & 49.24 & 6.23   
    & & 0.47 & 0.59
    \\
    \midrule
    \textbf{Ours}              
    & \textbf{-6.94} & \textbf{56.50} & \textbf{7.95} 
    & & \textbf{-7.30} & \underline{60.70} & \textbf{12.94}
    & & \textbf{-8.06} & \textbf{60.78} & \textbf{10.78}
    & & \textbf{0.54} & \textbf{0.67}
    \\
    \bottomrule
    \end{tabular}
    \begin{flushleft}
    {\footnotesize \hspace*{1em} Note: Evina $>$ 0 are considered invalid and are represented by “--”. Baseline ligand results are from CBGbench~\cite{Cbgbench}.}
    \end{flushleft}

    \caption{Comparative performance of baseline models and our method on binding affinity and drug-likeness metrics.}
    \label{tab:commands}

\end{table*}

\subsubsection{Global Attention with Adaptive Edge Selection.}

To capture the global protein pocket context that influences ligand topology and scaffold formation, we apply a unified attention mechanism over ligand atoms and virtual atoms.

Given features and coordinates \( \{ \mathbf{h}_a, \mathbf{x}_a \}_{a \in \mathcal{V}} \), where \(\mathcal{V}\) denotes the union of ligand atoms and virtual atoms, each node updates its representation by attending to all others:

\begin{equation}
\mathbf{h}_a^{(l+1)} = \mathbf{h}_a^{(l)} + \sum_{v \in \mathcal{V}, v \ne a}\psi_h\left( \mathbf{h}_a^{(l)}, \mathbf{h}_v, \mathbf{e}_{av}, \boldsymbol{\phi_{av}}\right)
\end{equation}

where \( \psi_h(\cdot) \) is a graph attention layer and \( \boldsymbol{\phi}_{av} \) encodes the inter-node distance.

To incorporate geometric influence, node coordinates are refined through a second attention stream:
\begin{equation}
\begin{aligned}
\mathbf{x}_a^{(l+1)} = \mathbf{x}_a^{(l)} 
+ \sum_{v \in \mathcal{V}, v \ne a} &\; (\mathbf{x}_a^{(l)} - \mathbf{x}_v^{(l)}) \cdot \\
&\hspace{-6em} \psi_x\left( \mathbf{h}_a^{(l+1)}, \mathbf{h}_v^{(l+1)},\mathbf{e}_{av}, \boldsymbol{\phi}_{av}  \right),
\end{aligned}
\end{equation}
where \( \psi_x \) is computed similarly to \( \psi_{h} \) but with separate parameters. This dual-stream mechanism allows the model to separately capture feature-level and geometry-level interactions.

To suppress weak or noisy interactions, we apply an adaptive edge selection strategy, retaining only edges whose mean attention score exceeds a threshold \( \tau \):
\begin{equation}
\mathcal{E}' = \left\{ (a, v) \in \mathcal{E} \;\middle|\; \frac{1}{H} \sum_{h=1}^{H} \alpha_{av}^{(h)} > \tau \right\}.
\end{equation}
\subsubsection{Local Interaction Refinement with Distance-Aware Edges.}
We designed fine-grained edge connection rules grounded in established domain knowledge, specifically inspired by the affinity calculations in AutoDock Vina~\cite{AutoDock}. This approach ensures that our modeling of molecular interactions is both accurate and biologically meaningful. Specifically, leveraging the interaction formulas from VinaDock, three types of edge with distance thresholds set to 2.7 \AA{}, 3.4 \AA{}, and 5 \AA{} are established. The first two thresholds are intended to capture steric and short-range interactions, while the 5 \AA{} threshold is specifically tailored to describe long-range interactions, such as van der Waals forces. These thresholds ensure that the model accurately captures different types of interactions based on spatial proximity, enhancing the precision of ligand generation conditioned on protein structures.
We apply the same graph attention modules as in the global stage, but restrict attention to explicitly connected node pairs, enabling fine-grained interaction modeling between ligand atoms and their spatially proximal protein neighbors.

\section{Experiments}

\subsection{Experimental Setup}

\subsubsection{Dataset.} We use the CrossDocked~\cite{CrossDocked} dataset, comprising 22.5 million docked protein–ligand complexes. Following standard protocols~\cite{luo20213d,Cbgbench,molcraft}, we select samples with Root-Mean-Square Deviation (RMSD) $<$ 1~\AA{} and sequence identity $<$ 30\%. This yields 100,000 complexes for training and 100 for testing. For each test complex, 100 ligands are generated to ensure robust evaluation.

\subsubsection{Metrics.}  
We evaluate model performance across four categories. \(\uparrow\) indicates higher is better, and \(\downarrow\) indicates lower is better.

\textit{Binding Affinity.}  
Following prior works, we use \textsc{Auto\-Dock Vina}~\cite{AutoDock} to evaluate binding energy from three perspectives: (a) Vina Score, which measures the energy of the original docked pose; (b) Vina Min, referring to the minimum energy after local minimization; and (c) Vina Dock, which represents the energy obtained through global redocking. Additionally, we compute \textbf{Evina}~($\downarrow$), the mean binding energy across all generated ligands; \textbf{IMP\%}~($\uparrow$), the percentage of generated ligands outperforming the reference; and \textbf{MPBG\%}~($\uparrow$), the average binding energy improvement over the reference.

\textit{Drug-likeness.}  
We assess drug-likeness using two standard metrics: Quantitative Estimate of Drug-likeness \textbf{(QED)}~($\uparrow$), which reflects oral bioavailability, and Synthetic Accessibility \textbf{(SA)}~($\uparrow$), which estimates synthetic feasibility.

\textit{Structural Plausibility.}  
Following Guan et al.~\cite{targetdiff}, we compare the distributions of bond lengths, C–C distances within 2\,\AA{}, and all-atom distances within 12\,\AA{}. Bond length categories include C--C, C=C, C:C, C--N, C=N, C:N, C--O, and C=O, where the symbols ``--'', ``='', and ``:'' respectively denote single, double, and aromatic bonds. These distributions are quantified using Jensen--Shannon Divergence, reported as \textbf{JSD\_BL}~($\downarrow$), \textbf{JSD\_CC\_2\AA}~($\downarrow$), and \textbf{JSD\_ALL\_12\AA}~($\downarrow$), where lower values indicate better alignment with empirical statistics.

\textit{Conformational Stability.}  
Following Harris et al.~\cite{posecheck}, we assess the conformational stability of generated ligands using two metrics: \textbf{Strain Energy (SE)}~($\downarrow$), which quantifies the internal energetic stability of the ligand and is reported at the 25th, 50th, and 75th percentiles, and \textbf{Steric Clashes (Clash)}~($\downarrow$), which measures the number of atomic overlaps between ligand and protein, reflecting spatial feasibility within the binding pocket.

\subsubsection{Baselines.}
We compare our method against a diverse set of baseline models, including both auto-regressive and non-auto-regressive approaches. Detailed descriptions of these baselines are provided in the Appendix.
\subsection{Results}

The performance of our method is evaluated across multiple metrics to comprehensively assess its effectiveness in generating high-quality ligands for SBDD. Results are shown in Tables~\ref{tab:commands}, \ref{tab:jsd_comparison}, and~\ref{tab:conformational_plausibility}, with the best scores in bold and the second-best underlined. Our method consistently outperforms baseline models across all dimensions, including binding affinity, structural plausibility, and conformational stability.

\textit{Binding Affinity and Drug-Likeness.}  
As shown in Table~\ref{tab:commands}, our method achieves the best Vina Score of -6.94, with 56.50\% of generated ligands outperforming reference ligands. In terms of MPBG\%, our method is the only one yielding a positive score 7.95\%, indicating overall improved binding affinity without any post-processing. For Vina Min and Vina Dock, SculptDrug also achieves optimal values of -7.30 and -8.06, respectively, demonstrating superior local energy minima and docking robustness. In terms of drug-likeness, SculptDrug yields competitive QED (0.54) and SA (0.67) scores, reflecting its potential to generate compounds with acceptable pharmacological and synthetic properties.

\textit{Structural Plausibility.}  
Table~\ref{tab:jsd_comparison} summarizes the Jensen–Shannon Divergence (JSD) scores for ligand structural distributions. 
SculptDrug consistently achieves the lowest divergence across all metrics: 
0.1522 for average bond lengths (JSD\_BL), 
0.1163 for short-range C–C bonds within 2~\AA{} (JSD\_CC\_2\AA), 
and 0.0331 for all-atom distances within 12~\AA{} (JSD\_All\_12\AA).  These results demonstrate that SculptDrug generates chemically valid structures with high fidelity to empirical spatial patterns, capturing local bonding accuracy.

\textit{Conformational Stability.}  
As shown in Table~\ref{tab:conformational_plausibility}, SculptDrug generates ligand conformations with reduced strain energy and fewer steric clashes, indicating improved spatial compatibility with the binding pocket and enhanced structural plausibility.
\begin{table}[t]
    \centering
   
    \setlength{\tabcolsep}{4pt}
    
    \begin{tabular}{@{}lccc@{}} 
    \toprule
     & \textbf{JSD\_BL} & \textbf{JSD\_CC\_2\AA} & \textbf{JSD\_All\_12\AA} \\ 
    \midrule
    GRAPHBP          & 0.5316           & 0.5229               & 0.3936                \\ 
    POCKET2MOL       & 0.5602           & 0.5236               & 0.2364                \\ 
    TARGETDIFF       & 0.2409           & 0.2285               & 0.0600                \\ 
    FLAG             & 0.4003           & 0.3611               & 0.0676                \\ 
    D3FG             & 0.3894           & 0.3416               & 0.1119                \\ 
    DECOMPDIFF       & 0.2277           & 0.1895              & 0.0486                \\ 
    DIFFSBDD         & 0.4610           & 0.4553               & 0.1614                \\ 
    MOLCRAFT         & \underline{0.2251} & \underline{0.1737}   & \underline{0.0417}    \\
    DIFFBP           & 0.5904           & 0.5926               & 0.3364                \\ 
    \cmidrule{1-4} 
    \textbf{Ours}    & \textbf{0.1522}  & \textbf{0.1163}      & \textbf{0.0331}       \\ 
    \bottomrule
    \end{tabular}
     \caption{Comparative analysis of ligand bond length distribution across generative models. }
     \label{tab:jsd_comparison}
\end{table}

\begin{table}[t]
    \centering
    \setlength{\tabcolsep}{5pt}
    \begin{tabular}{@{}lcccc@{}}
    \toprule
      & \textbf{SE\_25} & \textbf{SE\_50} & \textbf{SE\_75} & \textbf{Clash} \\ 
    \midrule
    GRAPHBP          & -                 & -                  & -                  & 193.67              \\ 
    POCKET2MOL       & -        & -        & -         & \underline{8.13}         \\ 
    TARGETDIFF       & 252.02            & 878.72             & -            & 9.33          \\ 
    FLAG             & 129.11                 & 353.60                  & 885.80                  & 42.59              \\ 
    D3FG             & 460.07           & 1332.99           & -              & 29.01         \\ 
    DECOMPDIFF       & 132.37            & 473.16           & 1904.26            & 8.51         \\ 
    DIFFSBDD         &895.07                 &-                  & -        & 71.18              \\ 
    MOLCRAFT         & \underline{83.92}            & \underline{197.10}            & \underline{536.22}             & 7.03         \\ 
    DIFFBP           & -                 & -                  & -                  & 78.17             \\ 
    \cmidrule{1-5} 
    \textbf{Ours}    & \textbf{72.90}   & \textbf{170.17}    & \textbf{523.66}   & \textbf{6.41}  \\ 
    \bottomrule
    \end{tabular}
    \begin{flushleft}
        {\footnotesize \hspace*{1em}``-'' indicates the value exceeds 10,000.}
    \end{flushleft}
    \caption{Evaluation of strain energy~(SE) and steric clashes~(Clash) in ligand-protein complexes.}
    \label{tab:conformational_plausibility}

\end{table}

\subsection{Ablation Studies}
\begin{figure}[h]
  \centering
  \includegraphics[width=\linewidth]{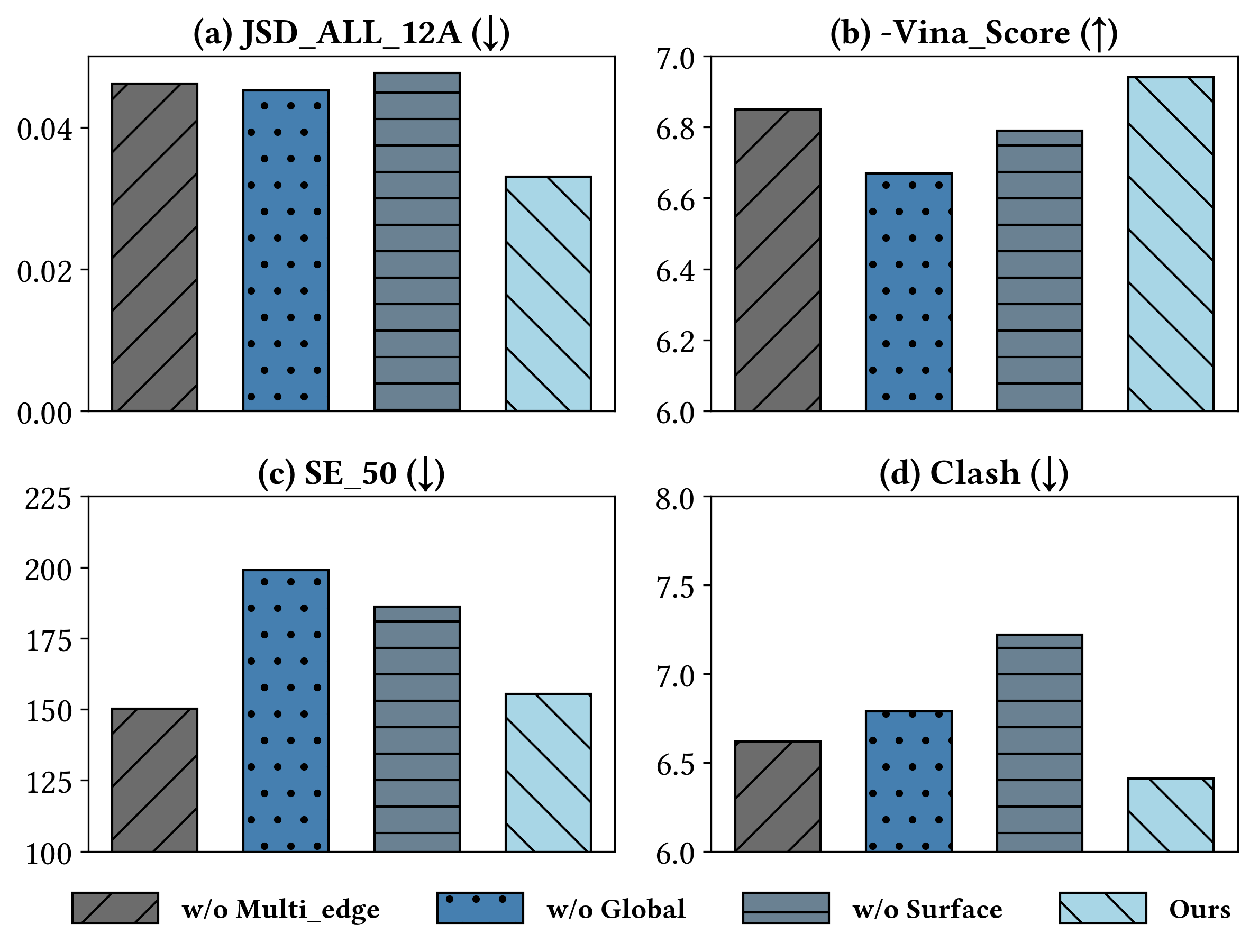}
  \caption{Impact of variants on SculptDrug 's performance.}
  \label{fig:ablation_results}
\end{figure}
To evaluate the contribution of each key component in SculptDrug , we conduct an ablation study by systematically removing individual modules: (1) \textbf{w/o Multi\_edge}, replacing the multi-type edge construction with a K-nearest neighbor (KNN) scheme; (2) \textbf{w/o Global}, which removes the global attention layer responsible for integrating global structural information through virtual atoms; and (3) \textbf{w/o Surface}, which omits the Boundary Awareness Block that incorporates protein surface geometry into ligand generation. The complete \textbf{SculptDrug } model serves as the baseline. Figure~\ref{fig:ablation_results} summarizes the performance metrics for each variant alongside the complete model. We can observe that removing any single component leads to a consistent decrease in performance across all metrics. 

Notably, the \textbf{w/o Global} variant leads to the most pronounced degradation in \textbf{Vina Score}, reflecting impaired ligand–protein compatibility due to missing global structural context. The \textbf{w/o Surface} variant results in elevated steric clashes and a slight increase in all-atom JSD, suggesting that surface-aware conditioning plays an important role in guiding spatially plausible ligand placement. The \textbf{w/o Multi\_edge} variant experiences a moderate decline, demonstrating that multiple edge types are essential for capturing complex local interactions and enhancing spatial modeling fidelity. In contrast, the complete \textbf{SculptDrug} model achieves the best performance overall, validating the synergistic effect of integrating hierarchical structure, surface information, and fine-grained interaction modeling.

\section{Conclusion and Future Works}
In this study, we propose SculptDrug, a novel Bayesian Flow Network-based SBDD model that effectively addresses three key challenges in ligand generation: boundary condition constraints, hierarchical structural integration, and spatial modeling fidelity. By employing a progressive denoising strategy and two tailored components—the Boundary Awareness Block and the Hierarchical Encoder—our method integrates protein surface geometry and multi-level structural information. A comprehensive evaluation on CrossDocked2020 demonstrates that SculptDrug surpasses state-of-the-art models in binding affinity, drug-likeness, structural plausibility, and conformational stability. Nonetheless, we observe that a small portion of generated ligands still exhibit high strain energy, indicating room for improvement. In addition, future work may consider incorporating dynamic protein modeling to further enhance biological relevance.

\footnotesize
\section*{Acknowledgements}
This work was partially supported by National Natural Science Foundation of China (62472174), the Shanghai Frontiers Science Center of Molecule Intelligent Syntheses.

\normalsize
\bibliography{main}


\footnotesize
\section*{Supplementary Related Work}
\subsection*{Hierarchical Graph Modeling}
Hierarchical modeling has been extensively applied in molecular graph learning. HiMol\cite{HiMol} and HimGNN~\cite{himgnn} partition molecules into atom, motif, and graph levels, and perform hierarchical aggregation for property prediction. HiGLLM~\cite{HiGLLM} explores hierarchical graph representations in the context of large language models. ISA-PN~\cite{ISA-PN} improves interpretability through positive and negative attention over hierarchical structures, while HMSN-CAM~\cite{HMSN-CAM} employs hierarchical co-attention to enhance drug–drug interaction (DDI) prediction. However, these approaches operate solely within molecular graphs and rely on fixed, rule-based motif partitions. They are primarily designed for small molecules and are not well suited to the complex, multi-scale structural context encountered in protein-conditioned settings. In structure-based drug design (SBDD), protein structures involve more flexible and diverse hierarchies, including local residue clusters, functional regions, and long-range interactions that are difficult to capture with simple partitioning rules.

\subsection*{Surface-based Modeling}
Several prior works have leveraged protein surface features for learning molecular or protein representations. For example,~\cite{pretrain_surface} and~\cite{surface_vqmae} integrate surface geometry with sequence and structural information to obtain global protein embeddings, while~\cite{surface_cvpr} learns surface-based point cloud representations for protein-level prediction. However, these methods focus on representation learning and do not address molecular generation or protein--ligand interactions.

SurfGen~\cite{surfgen} is more closely related to our setting, as it generates molecules based on protein surfaces. However, it models the protein solely at the surface level without incorporating atomic-level structural detail. Moreover, its autoregressive generation paradigm leads to slower inference and may increase the risk of generating redundant structures.

\section*{Baselines Details.}
We compare our method against a diverse range of baseline models, covering both auto-regressive and non-auto\-regres\-sive approaches as follows. 

\textit*{Autoregressive.} Pocket2Mol~\cite{pocketmol} focuses on pocket features; GRAPHBP~\cite{Graphbp} employs graph-based ligand generation; and FLAG~\cite{Flag} uses fragment motifs. 

\textit*{Non-autoregressive.} We include diffusion-based models such as TARGETDIFF~\cite{targetdiff}, DIFFBP~\cite{Diffbp}, and DIFFSBDD~\cite{diffsbdd}; D3FG~\cite{D3fg}, which integrates fragment motifs; DECOMPDIFF~\cite{Decompdiff}, employing arm and scaffold clustering; and MOLCRAFT~\cite{molcraft}, which is based on Bayesian Flow Networks.
This comprehensive selection ensures a robust comparison across diverse generative methodologies.

\begin{figure}[t]
  \centering
  \includegraphics[width=\linewidth]{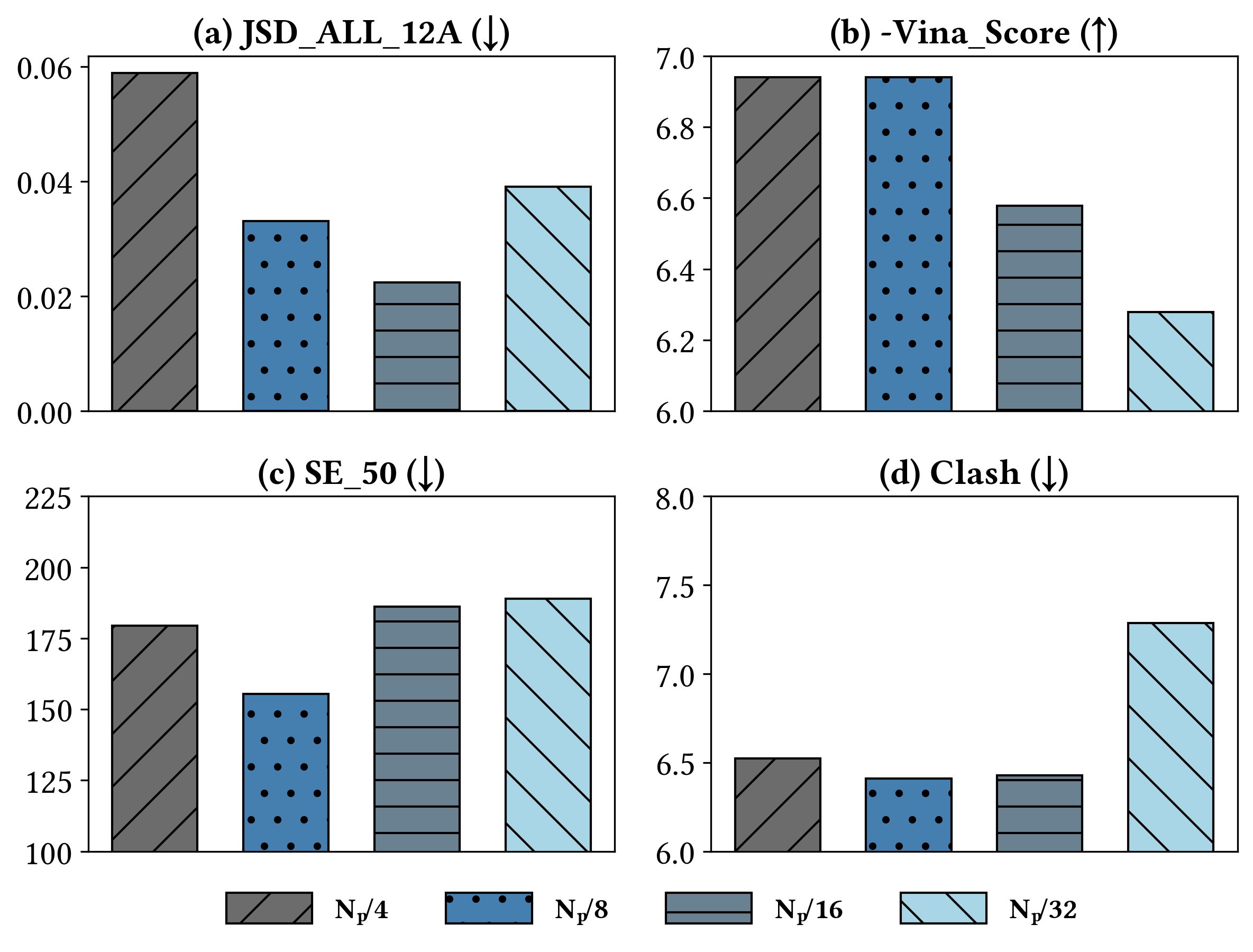}
  \caption{Impact of the number of clusters \(K\) on SculptDrug's performance.}
  \label{fig:parameter_results}
\end{figure}

\section*{Algorithm}

\begin{algorithm}[H]
\caption{Discrete-Time Loss Computation in SculptDrug}
\begin{algorithmic}[1]
\State \textbf{Input:}
\Statex \quad Atom positions $x_m \in \mathbb{R}^{3N_m}$, atom types $v_m \in \mathbb{R}^{N_m \times K}$
\Statex \quad Protein atoms $\mathcal{P} = \{ (\mathbf{x}_p^n, \mathbf{a}_p^n) \}_{n=1}^{N_p}$,
Surface $\mathcal{S} = \left( \left\{ (\mathbf{x}_s^n, \mathbf{a}_s^n) \right\}_{n=1}^{N_s}, \mathcal{E}_s \right)$
\Statex \quad Noise parameters: $\sigma_1$, $\beta_1$; Time steps $n$

\State Sample time step $i \sim \mathcal{U}(\{1, \dots, n\})$, set $t \gets \frac{i-1}{n}$
\State Coordinate accuracy schedule: $\beta_x(t) \gets \sigma_1^{-2} t - 1$
\State Type accuracy schedule: $\beta_v(t) \gets t^2 \cdot \beta_1$
\State Bayesian flow distribution:
\[
\mu \sim p_F^x(\mu \mid x_m, \mathcal{P}, \mathcal{S}; \beta_x(t))
\]
\[
\theta_v \sim p_F^v(\theta_v \mid v_m, \mathcal{P}, \mathcal{S}; \beta_v(t))
\]
\State Output distribution:
\[
\hat{x}, \hat{v} \gets p_O(\mu, \theta_v, \mathcal{P}, \mathcal{S}, t)
\]
\State Compute coordinate loss:
\begin{equation*}
\mathcal{L}_x^n \gets \frac{n(1 - \sigma_1^{2/n})^2}{2 \sigma_1^{2i/n}} \left\| x_m - \hat{x} \right\|^2
\end{equation*}
\State Compute precision $\alpha \gets \beta_1 \cdot \frac{2i - 1}{n^2}$
\State Sender distribution: $y_v \sim p_S^V(y_v \mid v_m; \alpha)$
\State Compute type loss:
\begin{equation*}
\mathcal{L}_v^n \gets \log p_S^v(y_v \mid v_m; \alpha) - \log p_R^V(y_v \mid \hat{v}; \alpha, t)
\end{equation*}
\State Compute total loss: $\mathcal{L}_n \gets \mathcal{L}_x^n + \mathcal{L}_v^n$
\State \Return $\mathcal{L}_n$
\end{algorithmic}
\end{algorithm}

\begin{algorithm}[htbp]
\caption{Sampling Procedure in SculptDrug}
\begin{algorithmic}[1]
\Require Protein input $\mathcal{P}$, surface $\mathcal{S}$, number of steps $N$, atom types $K$, Initial values: $\mu \leftarrow \mathbf{0}$, $\theta_v \leftarrow \left[\frac{1}{K}\right]^{n_M \times K}$, Noise parameters: $\sigma_1$, $\beta_1$
\For{$i = 1$ to $N$}
    \State $t \gets \frac{i-1}{N}$
    \State Output distribution:
    \[
    \hat{x}, \hat{v} \gets p_O(\mu, \theta_v, \mathcal{P}, \mathcal{S}, t)
    \]
    \State Update coordinate parameters:
    \[
    \beta(t) \gets \sigma_1^{-2} t - 1,
    \gamma \gets \frac{\beta(t)}{1 - \beta(t)}
    \]
    \[
    \mu \sim \mathcal{N}(\gamma \hat{x}, \gamma(1 - \gamma) \mathbf{I})
    \]
    \State Update type parameters:
    \[
    \alpha \gets \beta_1 \cdot \frac{2i - 1}{n^2}
    \]
    \[
    y_v \sim \mathcal{N}(\alpha(K \mathbf{e}_{\hat{v}} - 1),\alpha K \mathbf{I})
    \]
    \[
    \theta_v \gets \text{softmax}(y_v)
    \]
\EndFor
\State Final prediction:
\[
\hat{x}, \hat{v} \gets p_O(\mu, \theta_v, \mathcal{P}, \mathcal{S}, 1)
\]
\State \Return $\hat{x}, \hat{v}$
\end{algorithmic}
\end{algorithm}

\begin{figure*}[t]
  \centering
  \includegraphics[width=\textwidth]{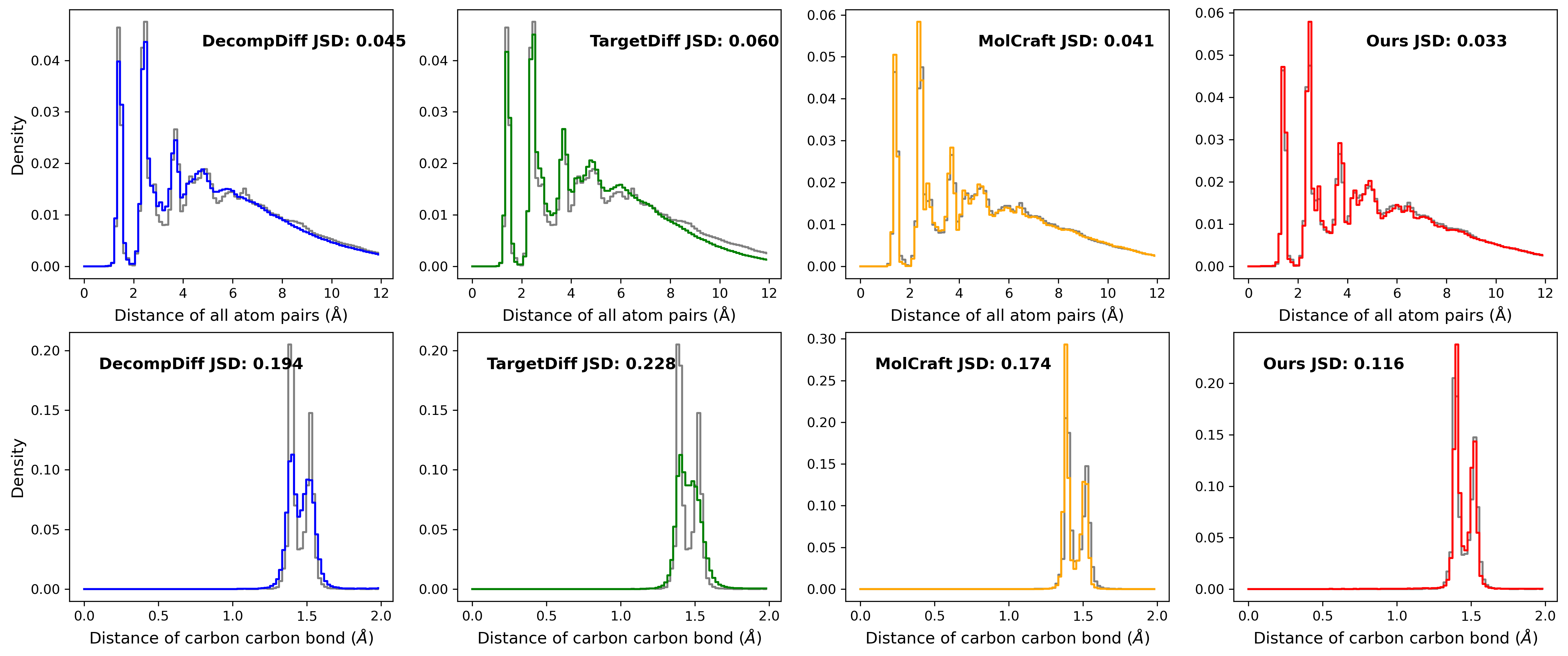}
  \caption{Comparison of distance distributions between generated ligands and empirical data(gray). Top row: distributions of all-atom distances within 12\,\AA{}; Bottom row: distributions of carbon–carbon bond lengths within 2\,\AA{}. }
  \label{fig:bond_distance_distribution}
\end{figure*}

\section*{EXPERIMENT DETAILS}
\subsection*{Model Parameter}
SculptDrug adopts a hierarchical architecture comprising three key modules: a two-layer Boundary Awareness Block, a two-layer Graph Attention Network (GAT) for global attention, and a nine-layer GAT for local interaction refinement. To enable global context modeling, we construct virtual atoms by applying k-means++ clustering to the 3D coordinates of protein pocket atoms. Based on hyperparameter analysis, we set the number of clusters to $N_p / 8$. Each GAT layer uses 16 attention heads, and all hidden dimensions are set to 128. The adaptive edge selection threshold is set to $\tau = 0.05$ to suppress weak or noisy interactions. 
\subsection*{Training Details}
In training SculptDrug, we adopt the core BFNs configurations from MolCraft\cite{molcraft}, utilizing 1,000 discrete time-steps to compute the discrete-time loss. The noise schedules are implemented with $\beta_1 = 1.5$ for atom types and $\sigma_1 = 0.03$ for atomic coordinate perturbations. The model is optimized over 26 epochs using Adam~\cite{Kingma2014AdamAM} with a learning rate of 0.005, momentum coefficients $(\beta_1, \beta_2) = (0.95, 0.999)$, and a batch size of 32. Additionally, we apply an exponential moving average (EMA) across all parameters with a decay rate of 0.999. All experiments are executed on an NVIDIA A800 GPU (80GB VRAM) with CUDA 12.4 acceleration.

\begin{figure}[t]
  \centering
  \includegraphics[width=\linewidth]{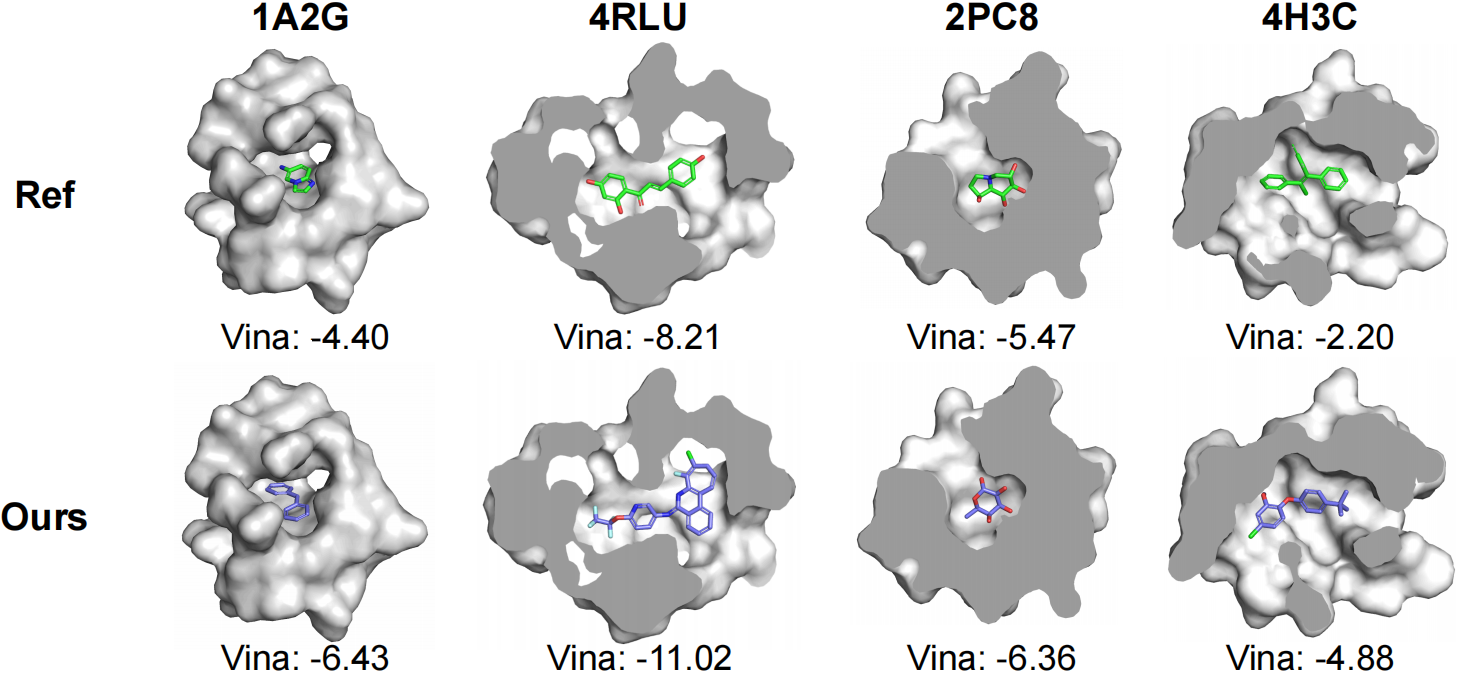}
  \caption{Visualization comparison between ligands generated by SculptDrug and reference ligands, selected based on the median Vina Score.}
  \label{fig:vis}
\end{figure}

\section*{Hyperparameter Analysis}

To systematically evaluate the impact of hyperparameter choices on molecular generation performance, we set the number of clusters in the k-means++~\cite{arthur2006k} algorithm to one of the following four values: \(N_p/4\), \(N_p/8\), \(N_p/16\), \(N_p/32\), where \(N_p\) denotes the total number of atoms in the protein binding pocket. Experimental results (Figure~\ref{fig:parameter_results}) show that setting \(K = N_p/8\) achieves an optimal balance between global structural modeling and local interaction refinement.
Specifically, when $K = N_p/4$, the JSD score increases significantly , suggesting that the large number of virtual atoms introduces excessive noise. In contrast, $K = N_p/32$ leads to a relatively poor Vina score and the highest clash score, indicating that the overly coarse granularity fails to capture crucial local structural details.  Our experiments demonstrate that choosing \(K = N_p/8\) preserves the essential local details while ensuring global structural consistency, thereby significantly enhancing ligand generation performance.
Therefore, we adopt \(K = N_p/8\) as the default configuration.

\section*{Examples of Generated Ligands}
Figure~\ref{fig:bond_distance_distribution} compares the distributions of all-atom distances (top row) and short-range carbon–carbon bond distances (bottom row) across four models. The gray histograms represent empirical distributions from reference data. Our method (rightmost column) exhibits the closest match to the reference.

In Figure~\ref{fig:vis}, we illustrate the ligands generated by SculptDrug alongside the reference ligands. We randomly select four protein pockets and present the ligands corresponding to the median Vina score. Notably, the generated ligands exhibit lower Vina Score, indicating stronger binding affinity.
\end{document}